\documentclass[letterpaper, 10 pt, conference]{ieeeconf}  %

\IEEEoverridecommandlockouts
\usepackage{amsmath} %
\usepackage{amssymb}  %

\usepackage{censor}

\usepackage{hyperref}
\usepackage{cleveref}

\usepackage{xcolor, colortbl, booktabs, multirow, graphicx, float, caption, cuted, tabularx, soul}

\newcommand{\methodname}{SCAGE}
\newcommand{\methodfullname}{SCene Anomaly Guided Exploration}

\newcommand{\networkname}{${\Delta}$-Net}

\definecolor{first}{RGB}{242, 182, 181}
\definecolor{second}{RGB}{249, 218, 184}
\definecolor{third}{RGB}{254, 250, 199}

\newcommand{\hlfirst}[1]{\sethlcolor{first}\hl{#1}}
\newcommand{\hlsecond}[1]{\sethlcolor{second}\hl{#1}}
\newcommand{\hlthird}[1]{\sethlcolor{third}\hl{#1}}

\hypersetup{
    colorlinks=true,
    linkcolor=blue,      %
    urlcolor=cyan,        %
}

\title{\LARGE \bf
Beyond Frontiers: Scene-Anomaly Guided Autonomous Exploration
}

\author{
Akash Kumbar$^1$ \quad
Abhinav Raundhal$^1$ \quad 
Madhava Krishna$^1$ \\
\thanks{$^{1}$Robotics Research Center, IIIT-Hyderabad, India}%
\thanks{$\dagger$Project page: \url{https://beyondfrontiers.github.io}}
}%

\IEEEaftertitletext{\vspace{-12mm}}

\begin{document}

\maketitle
\vspace{-12mm}

\thispagestyle{empty}
\pagestyle{empty}

\setlength{\stripsep}{1pt}
\begin{strip}
    \centering
        \includegraphics[width=0.95\linewidth]{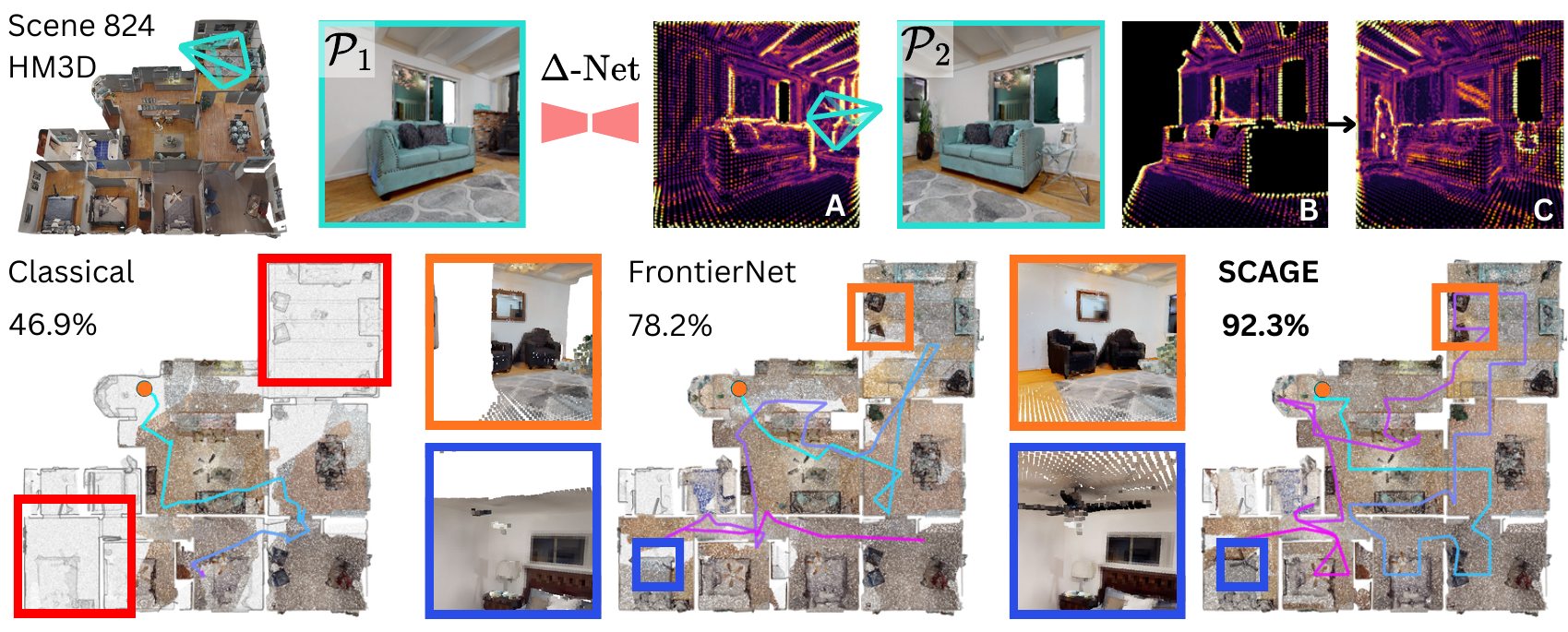}
        \captionof{figure}{
        \small{
        {\textbf{Top.} Our anomaly-guided exploration. Given a 3D point cloud at pose $\mathcal{P}_1$ (RGB images shown for context), our network \networkname{} predicts \textit{scene anomalies} (A) (shown in yellow on the point cloud), representing unexplored or poorly reconstructed areas.
        This anomaly signal guides the robot to an optimal next-best-view ($\mathcal{P}_2$). The anomalies as seen from $\mathcal{P}_2$, prior to obtaining the new point cloud, are shown in (B), which get resolved after obtaining the point cloud from this view as shown in (C). \textbf{Bottom.} Trajectory and mapping comparison on an HM3D scene from a starting position marked by an orange dot. Classical methods leave large sections unexplored (red boxes), and FrontierNet overlooks local geometric details. In contrast, our method, \methodname{} achieves the highest volumetric coverage (92.3\%) while simultaneously producing superior, high-fidelity 3D reconstructions (orange and blue insets highlight the reconstruction quality of fan and chair which FrontierNet fails to achieve).}
        }}
        \label{fig:teaser}
        \vspace{2mm}
\end{strip}

\begin{abstract}
Autonomous exploration of unknown 3D environments is traditionally driven by coverage-maximizing geometric heuristics. 
However, these methods typically determine exploration targets without considering the underlying structural context. This leads to inefficient trajectories often limiting the fidelity of the final 3D reconstruction. 
To bridge the gap between spatial coverage and reconstruction quality, we introduce a novel paradigm: reframing exploration as a geometric anomaly minimization problem. 
We present \methodname{}: \methodfullname{}, a novel autonomous exploration framework that operates directly on unstructured 3D point clouds. 
Instead of blindly chasing volumetric boundaries, we equip the robot with a foundational understanding of standard indoor architecture. 
As the robot navigates, it continuously evaluates its live 3D observations against these learned expectations. 
When the incoming geometry contradicts the learned priors of a typical indoor environment, such as a fragmented wall or a partial table, the system flags these regions as \textit{scene anomalies}. 
These geometric inconsistencies act as a guiding signal, naturally drawing the robot to investigate and resolve these structural anomalies from optimal vantage points. 
By actively targeting poorly reconstructed regions rather than just empty space, our approach seamlessly couples spatial discovery with high-fidelity mapping. 
Extensive evaluations demonstrate that \methodname{} achieves superior volumetric coverage ($\sim$90\% in all scenes) and higher 3D reconstruction quality compared to state-of-the-art baselines.
\end{abstract}

\section{INTRODUCTION}
\label{sec:introduction}

Autonomous exploration requires a mobile robot to navigate through unknown 3D environments to build digital maps, search for objects, or gather critical information. This capability is fundamental to a wide range of real-world applications, including search and rescue~\cite{tranzatto2022cerberus}, agricultural monitoring~\cite{gao2024aerial}, and infrastructure inspection~\cite{ginting2024semantic}.
Efficient autonomous exploration, whether aimed at maximizing mapped volume, enriching semantic understanding, or improving 3D reconstruction quality, ultimately boils down to determining the optimal sequence of poses.

Historically, exploration strategies have relied on geometric heuristics, either by driving the robot toward the boundaries between mapped and unmapped space~\cite{classicalfrontier,gao2018improved}, or by evaluating candidate poses against occupancy-derived metrics and selecting the most informative ones~\cite{bircher2016receding, schmid2020efficient}. In both cases, utility is computed by ray-casting into the occupancy grid and therefore reflects only the volume of unknown space a viewpoint reveals. Since unknown regions are undifferentiated and observed surfaces are considered resolved once marked occupied, these methods reward large open voids over occluded or sparsely sampled geometry, yielding trajectories that expand coverage while leaving structurally significant regions poorly reconstructed.

Recent learning-based methods attempt to overcome these limitations by using neural networks to predict the potential value of unmapped regions. While effective, these approaches typically rely on 2D visual cues~\cite{frontiernet, upadhyay2016fast} or external foundational models~\cite{tao2024learning}. More importantly, they maintain a narrow focus on maximizing spatial coverage.
This inadvertently decouples the act of exploring a space from the ultimate goal of generating a high-fidelity 3D reconstruction of it, leading to maps that may be complete in volume but lacking in geometric detail or structural integrity (as shown in Fig. ~\ref{fig:teaser}). While recent research has begun utilizing scene completion networks~\cite{schmid2022sc} and semantic priors~\cite{tao2022seer} to predict information gain through extrapolating unobserved volumes, these explicit hallucination methods often act as weak exploration indicators due to their limited capability to reliably complete highly complex indoor scenes.

In this paper, we introduce a novel paradigm in autonomous exploration: rather than simply searching for volumetric boundaries, we reframe exploration as a geometric anomaly-minimization problem. 
We ask: \textit{Which part of our current reconstruction fails to meet our learned architectural priors?} By targeting these geometric contradictions, we naturally guide the robot toward unresolved or occluded regions that are otherwise neglected.

Typical indoor environments exhibit predictable geometric distributions like flat walls, orthogonal intersections, and standard furniture layouts. Incomplete, sparse, or occluded observations inherently deviate from these distributions creating what we define as \textit{scene anomalies}, localized regions where the observed 3D geometry contradicts established architectural logic. We propose \methodname{}: \methodfullname{}, a novel exploration framework that operates directly on 3D point clouds to identify these anomalies. At the core of our framework is \networkname{}, a denoising autoencoder based on the Point Transformer V3~\cite{ptv3} architecture.
Designed to learn the geometric priors of typical indoor structures, \networkname{} explicitly predicts the required point-wise distribution shift (the $\Delta$) between a partial observation and its expected architectural form. Crucially, we train the network on localized point cloud chunks rather than entire spatial layouts, to focus on learning the fundamental structural features that compose typical indoor environments.

During online exploration, the network predicts a point-wise geometric deviation vector for incoming observations. Highly anomalous regions are clustered to compute adaptive 6-DoF observation viewpoints dynamically. Guided by a two-step lookahead utility planner, this anomaly-driven approach naturally steers the robot toward unexplored and under-reconstructed areas, allowing the system to maximize coverage and minimize reconstruction errors jointly.

In summary, the key contributions of our work are:
\begin{itemize}
\item We reframe 3D autonomous exploration as an anomaly minimization problem, introducing a novel framework, \methodname{}, that uses geometric deviation to drive navigation.

\item We design a denoising autoencoder \networkname{} that efficiently learns the structural priors of standard indoor environments from localized point cloud chunks.

\item We demonstrate through extensive evaluations that \methodname{} achieves superior volumetric coverage ($\sim$15\% improvement) and higher 3D reconstruction quality (half the error) as compared to state-of-the-art exploration baselines.
\end{itemize}

\section{RELATED WORK}
\label{sec:related_work}

\textbf{Geometry-Based Exploration.} Traditional exploration relies on geometric heuristics to identify optimal viewpoints. Frontier-based methods extract boundaries between known and unmapped regions to direct navigation \cite{classicalfrontier, gao2018improved}. Alternatively, sampling-based approaches evaluate candidate poses against metrics like map entropy and uncertainty to maximize volumetric coverage \cite{bircher2016receding, schmid2020efficient}. Hybrid approaches bridge these paradigms, combining the targeted efficiency of frontiers with the evaluative power of sampling. This is achieved either by evaluating frontier-sampled viewpoints via entropy and travel-time utility functions \cite{dai2020fast}, or by pairing global frontier exploration with local Next-Best-View Planning (NBVP) \cite{selin2019efficient}. Recent hierarchical frameworks further optimize these strategies for large-scale 3D environments \cite{cao2021tare}. Although these methods maximize mapped volume, they overlook the underlying structural context of the environment often causing myopic over-exploration of local regions at the expense of discovering new spaces.

\textbf{Learning-Based Exploration.} To overcome the limitations of purely geometric methods, recent literature increasingly leverages learning-based vision algorithms. For instance, several approaches estimate the information gain of candidate frontiers using 3D occupancy prediction models or scene completion networks \cite{schmid2022sc, tao2022seer}, while others bypass dense 3D maps entirely by predicting frontier utility directly from 2D visual cues \cite{frontiernet}. Emerging frameworks explore novel map structures, utilizing neural implicit representations \cite{yan2023active, lee2022uncertainty} and 3D Gaussians \cite{jiang2024ag, jiang2024fisherrf, tao2025rt}. Beyond pure geometry, incorporating appearance data such as object-level semantics into 3D representations has proven effective for refining trajectory and viewpoint evaluation \cite{papatheodorou2023finding, alama2025rayfronts, kim2025raven}. Finally, alternative paradigms treat exploration as a decision-making problem, employing reinforcement learning on color images \cite{tao2024learning, chaplot2018active, chaplot2021seal}, or utilizing vision foundation models and LLMs to guide human-like navigation \cite{yokoyama2024vlfm, chen2023not, qu2024ippon}.

\textbf{Redefining Frontiers for Exploration.} Unlike existing learning-based methods that rely on 2D visual cues or voxel-grid completions which focus on maximizing coverage, our approach operates directly on 3D point clouds. We shift the objective from simply finding empty space to detecting structural anomalies, guiding the robot toward areas that deviate from the expected geometry of typical indoor environments. This allows our system to jointly maximize coverage and minimize reconstruction errors by naturally targeting unresolved or occluded regions that standard volume-driven methods ignore.

\begin{figure*}[!t]
    \centering
    \includegraphics[width=0.99\linewidth]{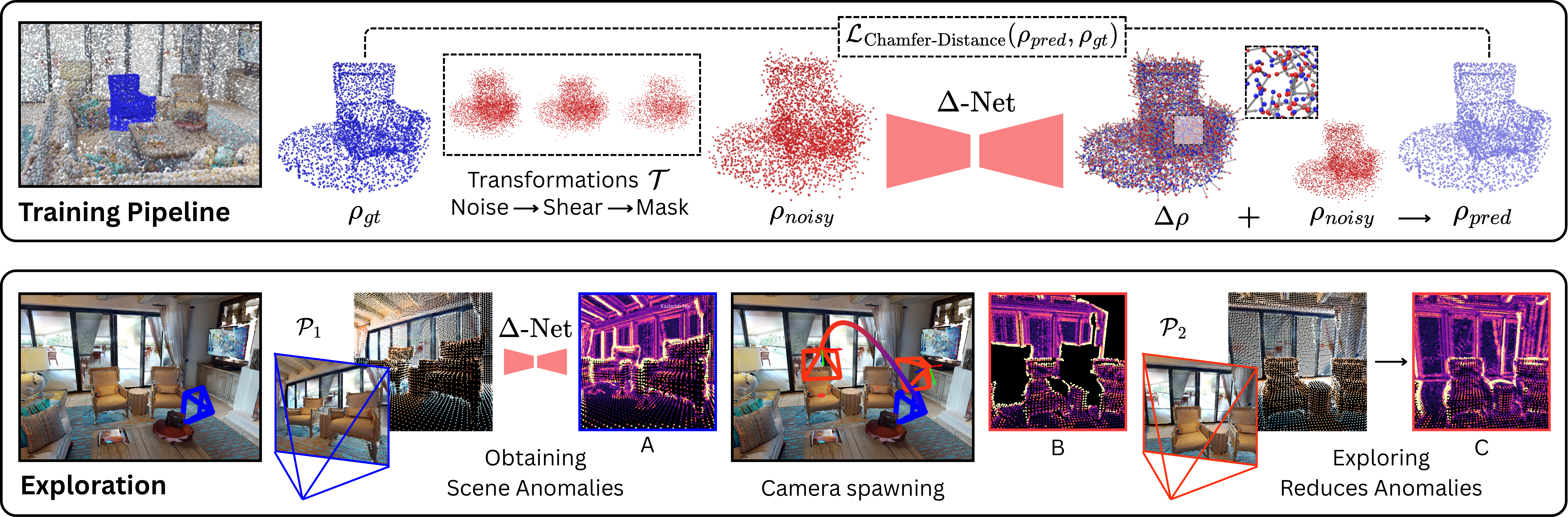}
    \caption{
    \small{
    \textbf{Overview of \methodname{}.} 
    \textbf{Top.} The Training Pipeline. Given point clouds of indoor scenes, we partition them to obtain ground truth chunks $\rho_{gt}$. We apply transformations $\mathcal{T}$ (e.g., Noise, Shearing, Masking) to simulate partial observability, yielding $\rho_{noisy}$. Our $\Delta$-Net architecture processes this corrupted input to predict a point-wise distribution shift $\Delta \rho$, visualized as per-point arrows mapping the noisy input $\rho_{noisy}$ toward the ground truth $\rho_{gt}$. The predicted $\Delta \rho$ is added to $\rho_{noisy}$ to obtain $\rho_{pred}$. The network is trained by minimizing the $\mathcal{L}_{\text{Chamfer-Distance}}$ between $\rho_{pred}$ and $\rho_{gt}$. 
    \textbf{Bottom.} The Exploration phase. Given a partial 3D observation from an initial pose $\mathcal{P}_1$, $\Delta$-Net predicts the distribution shift, which is thresholded to identify \textit{Scene Anomalies} (A), with yellow representing high-anomaly regions. Cameras are spawned to target these anomalies, generating candidate 6-DoF viewpoints (red frustums). The robot then navigates to the optimal pose $\mathcal{P}_2$. The anomalies, as seen from $\mathcal{P}_2$ (B), are resolved (C) once the new point cloud is captured, enabling high-fidelity 3D reconstruction.}
}
    \vspace{-2mm}
    \label{fig:method_fig}
\end{figure*}

\section{{\methodname} OVERVIEW}

We present an overview of the proposed \methodname{} pipeline in Fig.~\ref{fig:method_fig}. The framework consists of two primary phases. First, in the training pipeline (Section ~\ref{training_pipeline}), we train a point-based denoising autoencoder to learn the underlying structural priors of typical indoor environments. Second, during the exploration phase (Section ~\ref{inference_pipeline}), we leverage the point-wise distribution shift from these learned priors to identify unseen regions. The magnitude of these structural deviations serves as an anomaly score to drive autonomous exploration.

\subsection{Learning what indoor scenes look like}
\label{training_pipeline}

\textbf{Data Preparation.} Our objective is to capture the geometric distribution of complete, fully observed indoor scenes. To achieve this, we partition 3D point clouds of various indoor scenes from the Matterport3d dataset~\cite{matterport3d} into localized chunks $\rho_{gt} \in \mathbb{R}^{N \times 3}$ ($N = 4096$). To construct supervised training pairs that teach the network to map degraded observations back to their idealized structural form, we apply a series of stochastic transformations $\mathcal{T}$ to these ground-truth chunks. Specifically, we corrupt the input by injecting Gaussian noise with varying standard deviations, applying anisotropic scaling, and utilizing random point masking (as illustrated in \cref{fig:method_fig}) to simulate geometric distortions, partial coverage, and occlusions that arise from real-world depth sensing, yielding noisy input chunks: $\rho_{noisy} = \mathcal{T}(\rho_{gt})$.

\textbf{Model Architecture and Training.} Our architecture is built upon the Point Transformer V3~\cite{ptv3} backbone, which efficiently processes 3D data, serializing unstructured 3D point sets into 1D sequences via space-filling curves and applying localized serialized attention. Given a noisy input point cloud chunk $\rho_{noisy} \in \mathbb{R}^{N \times 3}$, the proposed network, \networkname{} employs a hierarchical encoder-decoder structure, which we train from scratch. 

The encoder progressively downsamples the point resolution into a compact 512-channel latent representation. Crucially, we enforce a strict information bottleneck by excluding any skip connections between the encoder and decoder, preventing the network from bypassing this compressed representation and directly copying input features. This design choice forces the autoencoder to internalize the geometric priors of typical indoor structures rather than overfit to the specific corruptions present in the training data. The symmetric decoder then upsamples this latent representation back to the original point resolution, and a final Multi-Layer Perceptron (MLP) head predicts a point-wise distribution shift vector $\Delta \rho \in \mathbb{R}^{N \times 3}$. The final denoised coordinates are then computed as $\rho_{pred} = \rho_{noisy} + \Delta \rho$.

During training, \networkname{} is optimized to reconstruct the underlying geometry by minimizing the Chamfer Distance between the predicted chunk $\rho_{pred}$ and the ground truth chunk $\rho_{gt}$. 

{\footnotesize
\begin{equation}
\mathcal{L}_{CD}(\rho_{pred}, \rho_{gt}) = \sum_{x \in \rho_{pred}} \min_{y \in \rho_{gt}} \|x - y\|_2^2 + \sum_{y \in \rho_{gt}} \min_{x \in \rho_{pred}} \|x - y\|_2^2
\end{equation}
}

This bidirectional loss penalizes spatial divergence, forcing the autoencoder to map anomalous points back to the learned distribution of standard indoor architecture. 
The resulting point-wise distribution shift vector $\Delta \rho$ captures how strongly each observed region deviates from the learned geometric prior, and is leveraged during inference to guide the robot toward structurally incomplete areas, as described next.

\begin{figure}[t]
    \centering
    \includegraphics[width=1\linewidth]{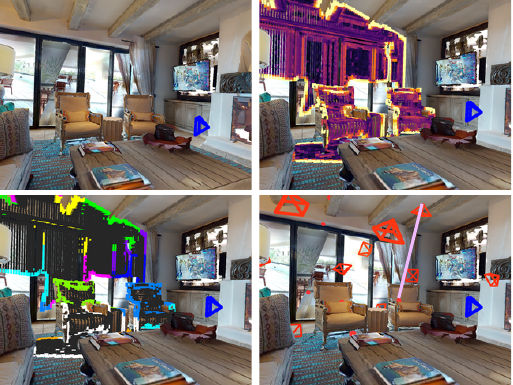}
    \caption{
    \small{
    \textbf{Viewpoint Generation from Scene Anomalies}. Top Left: A part of a scene with a camera shown with a blue frustum. Top Right: Per-point anomaly scores obtained from this camera view, with high-anomaly regions shown in yellow. 
    Bottom Left: Anomalies are clustered into regions shown in different colors.
    Bottom Right: For each cluster, the cluster anchor is shown with a red dot and the corresponding candidate 6-DoF viewpoints (red frustums) are spawned facing each anomaly cluster.
    They are placed at a standoff distance from the cluster anchor (shown with a pink line for one of the cameras), adapting to the spatial extent of the cluster. Larger clusters (e.g., the wall boundary) produce viewpoints at greater standoff distances, while compact clusters are observed from nearby, ensuring each anomalous region is fully captured within the camera's field of view.}
    }
    \label{fig:cam_spawn}
    \vspace{-5mm}
\end{figure}

\subsection{Exploration using Scene Anomalies}
\label{inference_pipeline}

\textbf{Detecting Scene Anomalies.} During exploration, starting from the initial camera pose, we obtain a depth map and backproject it into 3D to obtain a point cloud of the observed scene. 
In subsequent iterations, the newly backprojected points are merged with the accumulated scene representation to form a global point set $P$.

To maintain uniform spatial density and control computational cost, we apply voxel downsampling with resolution $0.05$m to obtain a filtered point cloud $\tilde{P}$. 
The downsampled point cloud is then partitioned into overlapping localized chunks, each processed independently by \networkname{}.

For each chunk $\rho \in \mathbb{R}^{N \times 3}$, the network predicts a point-wise distribution shift
$
\Delta \rho \in \mathbb{R}^{N \times 3}.
$
For a point $p_i \in \rho$, let its predicted deviation vector be $\Delta p_i$.
If a point belongs to multiple overlapping chunks, we aggregate its deviation magnitude by taking the maximum response. This produces a global deviation field over $\tilde{P}$.

We then compute the globally normalized anomaly score
\begin{equation}
s_i =
\frac{\|\Delta p_i\|_2 - \min_j \|\Delta p_j\|_2}
{\max_j \|\Delta p_j\|_2 - \min_j \|\Delta p_j\|_2},
\quad s_i \in [0,1]
\end{equation}
where the extrema are evaluated over all points in $\tilde{P}$.

These per-point anomaly scores quantify the degree to which local geometry deviates from the learned structural prior and serve as the signal for exploration. To reduce redundant computation, we skip re-processing points lying in low-anomaly neighborhoods, thereby concentrating inference on structurally uncertain and unexplored regions.

\textbf{Explore the unseen.} To convert these anomaly signals into navigable targets, we extract the anomalous set
$
\mathcal{A} = 
\{ p_i \in \tilde{P} \mid s_i > \tau \},
$
where $\tau = 0.5$ is an empirically selected threshold.
Because these anomalous points often represent contiguous unobserved regions or boundaries, we group $\mathcal{A}$ into distinct target clusters $C_k$, using ball-query based clustering.

\textbf{Viewpoint Generation.} For each cluster $C_k$, we compute a 6-DoF observation viewpoint that directly addresses the detected anomaly (as illustrated in Fig.~\ref{fig:cam_spawn}). We first obtain the geometric center (anchor) $a_k$ and extract the local surface normal $\hat{n}_k$ toward the sensor origin to ensure the resulting viewpoint lies in observed free space.

The observation distance must adapt to the spatial extent of each anomaly: small clusters can be observed from nearby, while spatially extended regions require the camera to retreat so the full cluster remains visible. Concretely, we compute $d_\text{standoff}$, the minimum distance at which a cluster of spatial radius $r_k = max_{p_j \in C_k} \|p_j - a_k\|_2$ subtends an angle no greater than the camera's horizontal field of view:
\begin{equation}
    d_{\text{standoff}} = \max\left(d_{\min}, \frac{r_k}{\tan(\theta_{\text{FOV}}/2)}\right)
    \label{eq:standoff}
\end{equation}

where $d_{min}$ enforces a physical safety margin. The candidate camera position is then $\mathcal{P}_{cam,k} = a_k + d_\text{standoff} \cdot \hat{n}_k$, with the camera yaw oriented toward $a_k$, fully defining a preliminary 6-DoF target pose $T_k$.

\textbf{Target Refinement and Selection}. Because dense anomaly clustering can produce redundant or kinematically unreachable targets, all candidates undergo a rigorous refinement phase. Targets are spatially consolidated to merge highly overlapping views, followed by free-space and line-of-sight validation against a volumetric occupancy grid. Then, for each surviving candidate $T_k$, we evaluate its exploratory value using the following utility function:
\begin{equation}
u(T_k) = \frac{g(T_k) \cdot \overline{s}_k}{\max(\|\mathcal{P}_{\text{robot}} - \mathcal{P}_{\text{cam},k}\|_2, \epsilon)}
\label{eq:utility}
\end{equation}

where $g(T_k)$ is the expected volumetric information gain computed via raycasting, $\overline{s}_k$ is the mean anomaly score of cluster $C_k$, and the denominator represents the Euclidean traversal cost from the robot's current position. This formulation balances three competing objectives: preferring views that reveal new volume, that resolve high-anomaly regions, and that minimize travel distance. We ablate this design in Table~\ref{tab:ablation_utility}, demonstrating all three terms are necessary for strong performance. The robot navigates to the utility-maximizing target via a collision-free target path planned using the RRT* algorithm~\cite{rrt*}.

\section{EXPERIMENTS AND RESULTS}

\begin{figure*}[!t]
    \centering
    {\small
    \begin{tabularx}{\linewidth}{*{4}{>{\centering\arraybackslash}X}}
    Classical~\cite{classicalfrontier} & NBVP~\cite{schmid2020efficient} & FrontierNet~\cite{frontiernet} & \textbf{\methodname{}} \\
    \end{tabularx}}
    \includegraphics[width=0.98\linewidth]{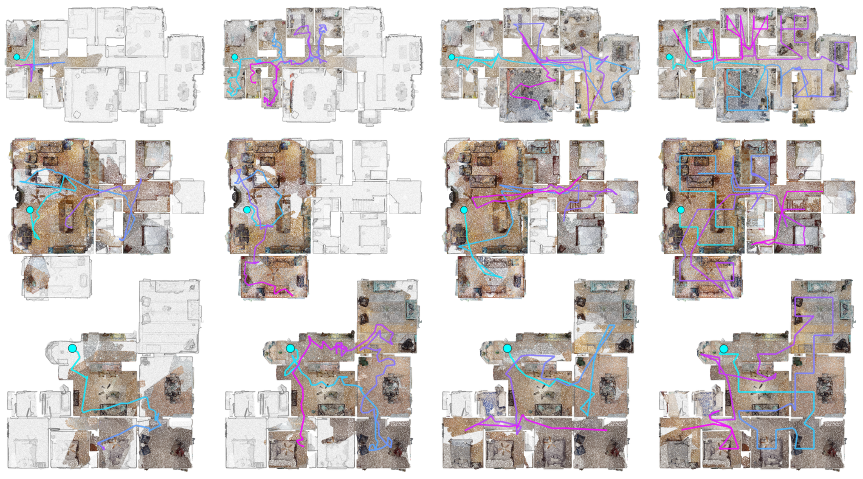}
    \caption{
    \small{
    \textbf{Qualitative Comparison. (Coverage and Trajectory).} Exploration results of our method compared against baselines on 3 scenes (top to bottom: 804, 812, 824). Starting location is marked with a cyan point, unexplored regions are shown in grey. The trajectory is shown in a gradient from cyan at the start to pink towards the end. Our method, \methodname{}, achieves the maximum coverage across all baselines on all scenes. Though FrontierNet~\cite{frontiernet} seems to achieve high coverage, it often just visits the room at its entrance, failing to capture structural elements that compose the room like chairs or beds. Whereas, by modeling the exploration as an anomaly minimization problem, we achieve both maximum coverage and good quality reconstruction, capturing all the fine details that construct indoor scenes.}}
    \label{fig:results_coverage}
    \vspace{-5mm}
\end{figure*}

\subsection{Experimental Setup}

\textbf{Implementation Details}. In order to ensure that \networkname{} learns geometric priors from structurally complete environments, we curate a training set of 75 high-fidelity scenes from the Matterport3D dataset. These scenes were specifically selected for their high structural integrity, actively filtering out scans with significant artifacts or missing geometry. By training \networkname{} exclusively on these clean, continuous manifolds from Matterport3D, and evaluating on unseen HM3D validation scenes, we demonstrate the robust cross-dataset generalization of our learned prior. The network was trained for 600 epochs on a single NVIDIA RTX A6000 GPU. 

\textbf{Benchmark Standardization and Simulation}. As recently highlighted by FrontierNet~\cite{frontiernet}, the evaluation of autonomous exploration systems is frequently hindered by the lack of standardized test protocols and a reliance on limited, disparate scenarios. Concurring with this assessment, we adopt their proposed evaluation framework to ensure a transparent comparison. We evaluate our approach using the identical subset of HM3D validation scenes, spanning diverse geometric complexities and spatial scales, and enforce the same maximum number of simulation steps per episode. By freezing the validation split, episode horizons, and sensor specifications, we isolate our learned priors as the sole source of performance gains. We simulate camera viewpoints and render images using Open3D, generating depth maps at a resolution of $480 \times 480$. The sensor is configured with a field of view (FOV) of $77.32^\circ \times 77.32^\circ $ and a maximum depth range of 5m.
Depth input is processed in two variants: (1) perfect depth rendered directly from the simulation and (2) noisy depth predicted by Metric3D v2~\cite{metric3dv2}. We maintain a volumetric occupancy map using Octomap and rely on the Open Motion Planning Library (OMPL) for low-level 3D path generation.

\subsection{Results}
\label{results4b}

\begin{table*}[]
\centering
\small
\setlength{\tabcolsep}{2pt}
\renewcommand{\arraystretch}{1.1}
\resizebox{0.98\linewidth}{!}{%
\begin{tabular}{c l  cccccccccc  c}
\toprule
 & & \multicolumn{10}{c}{Scene ID} \\
\cmidrule(lr){3-12}
  & Method &  804 & 807 & 812 & 824 & 827 & 834 & 854 & 876 & 879 & 880 & Mean \\ \midrule

\multirow{5}{*}{Volume Coverage $\uparrow$}
& Classic~\cite{classicalfrontier} & $30.9 \pm 5.3$& $45.6 \pm 2.2$ & $64.5 \pm 4.8$ & $48.9 \pm 2.3$ & $59.1 \pm 7.9$ & $49.7 \pm 5.3$ & $51.3 \pm 4.4$ & $44.2 \pm 8.8$ & $50.2 \pm 4.1$ & $62.4 \pm 5.3$ & $50.7$ \\
& NBVP~\cite{schmid2020efficient} & $37.1 \pm 6.7$ & $50.1 \pm 3.2$ & \cellcolor{third}$83.7 \pm 4.1$ & $64.3 \pm 4.8$ & \cellcolor{third}$76.7 \pm 4.1$ & $64.0 \pm 9.4$ & $80.3 \pm 2.3$ & $61.4 \pm 10.3$ & $63.1 \pm 8.4$ & $47.2 \pm 3.1$ & $62.8$ \\
& FrontierNet~\cite{frontiernet} & \cellcolor{third}$55.4\pm 7.1$& \cellcolor{third}$56.1\pm 6.1$& $81.8\pm 2.2$& \cellcolor{third}$72.4\pm 6.2$& $73.4\pm 9.1$& \cellcolor{third}$74.5 \pm 10.3$& \cellcolor{first}$93.1\pm 5.3$& \cellcolor{third}$75.9 \pm 8.5$& \cellcolor{third}$70.8\pm 10.1$& \cellcolor{third}$69.2 \pm 10.1$& \cellcolor{third}$72.3$ \\
& \textbf{\methodname{}} & \cellcolor{first}$92.5 \pm 1.0$& \cellcolor{first}$84.5 \pm 0.3$& \cellcolor{first}$88.8 \pm 0.5$& \cellcolor{first}$93.0 \pm 0.5$& \cellcolor{first}$90.0 \pm 0.5$& \cellcolor{first}$84.1 \pm 0.5$& \cellcolor{third}$91.4 \pm 3.5$& \cellcolor{second}$91.0 \pm 0.1$& \cellcolor{first}$90.4 \pm 0.1$& \cellcolor{first}$91.6 \pm 0.8$& \cellcolor{first}$89.7$ \\
& \textbf{\methodname{}*} & \cellcolor{second}$90.1 \pm 1.9$& \cellcolor{second}$81.5 \pm 2.4$& \cellcolor{second}$85.8 \pm 1.8$& \cellcolor{second}$90.0 \pm 2.1$& \cellcolor{second}$87.3 \pm 3.5$& \cellcolor{second}$83.1 \pm 1.1$& \cellcolor{second}$92.1 \pm 4.5$& \cellcolor{first}$92.9 \pm 1.6$& \cellcolor{second}$88.4 \pm 0.3$& \cellcolor{second}$90.1 \pm 3.8$& \cellcolor{second}$88.1$ \\
\midrule

\multirow{5}{*}{Chamfer($\times 10^{-3}$) $\downarrow$}
& Classic~\cite{classicalfrontier} & $12.0 \pm 4.0$ & $11.0 \pm 3.5$ & $3.5 \pm 1.0$ & $5.5 \pm 2.0$ & $6.0 \pm 2.5$ & $4.5 \pm 2.0$ & $1.5 \pm 1.0$ & $4.5 \pm 2.0$ & $6.5 \pm 2.5$ & $10.5 \pm 4.0$ & $6.55$ \\
& NBVP~\cite{schmid2020efficient} & $9.5 \pm 3.5$ & $8.5 \pm 3.0$ & $1.8 \pm 0.8$ & $3.5 \pm 1.5$ & $3.8 \pm 1.8$ & $2.8 \pm 1.5$ & $0.8 \pm 0.5$ & $2.8 \pm 1.2$ & $4.5 \pm 2.0$ & $8.0 \pm 3.5$ & $4.60$ \\
& FrontierNet~\cite{frontiernet} & \cellcolor{third}$7.1 \pm 3.4$ & \cellcolor{third}$6.5 \pm 2.9$ & \cellcolor{second}$0.9 \pm 0.3$ & \cellcolor{second}$1.9 \pm 1.4$ & \cellcolor{third}$2.1 \pm 1.5$ & \cellcolor{first}$1.4 \pm 0.9$& \cellcolor{first}$0.6 \pm 0.2$ & \cellcolor{third}$1.5 \pm 1.2$& \cellcolor{third}$2.9 \pm 1.8$& \cellcolor{third}$5.3 \pm 4.3$& \cellcolor{third}$3.02$ \\
& \textbf{\methodname{}} & \cellcolor{first}$1.2\pm0.8$& \cellcolor{first}$0.5\pm0.3$ & \cellcolor{first}$0.7\pm0.2$ & \cellcolor{first}$1.0\pm0.6$ & \cellcolor{first}$0.6\pm0.4$ & \cellcolor{second}$1.6\pm0.4$ & \cellcolor{second}$0.5 \pm 0.0$ & \cellcolor{first}$0.6\pm0.1$ & \cellcolor{first}$0.6\pm0.1$ & \cellcolor{first}$1.7\pm0.5$ & \cellcolor{first}$0.91$ \\
& \textbf{\methodname{}*} & \cellcolor{second}$2.4\pm1.2$& \cellcolor{second}$1.2\pm0.6$ & \cellcolor{third}$1.6\pm0.5$ & \cellcolor{third}$2.1\pm1.0$ & \cellcolor{second}$1.3\pm0.7$ & \cellcolor{third}$3.1\pm0.9$ & \cellcolor{third}$1.3 \pm 0.3$ & \cellcolor{second}$1.4\pm0.4$ & \cellcolor{second}$1.3\pm0.3$ & \cellcolor{third}$3.2\pm1.1$ & \cellcolor{second}$1.89$ \\
\bottomrule
\end{tabular}%
}
\caption{
\small{\textbf{Quantitative Comparison.} Results comparing Volumetric Coverage and Reconstruction Quality quantified by Chamfer Distance. Our method achieves the maximum coverage ($\sim$90\%, a substantial $\sim$15\% improvement on an average over the best baseline) in all scenes, showing its capability to explore without missing any parts of the scene. More importantly, along with the coverage, we achieve the best reconstruction quality, as shown by lowest chamfer distance. We obtain less than half the reconstruction error, on an average, as compared with the best baseline. Notice the low standard deviation of our method, highlighting its robustness towards different initial positions across multiple runs. The 3 digit numbers in the first row are scene IDs. Results in the first 4 rows are using depth from the simulator including \methodname{}, whereas the last row \methodname{}* uses the depth estimated by Metric3D v2~\cite{metric3dv2}. Red, orange, and yellow highlights indicate the \hlfirst{first}, \hlsecond{second}, and \hlthird{third} best performing technique.}}
\label{tab:quantitative_results}
\end{table*}

\begin{figure}[t]
    \centering
    {\small
    \begin{tabularx}{\linewidth}{*{4}{>{\centering\arraybackslash}X}}
    Classical~\cite{classicalfrontier} & NBVP~\cite{schmid2020efficient} & FrontierNet~\cite{frontiernet} & \textbf{\methodname{}} \\
    \end{tabularx}}
    \includegraphics[width=0.98\linewidth]{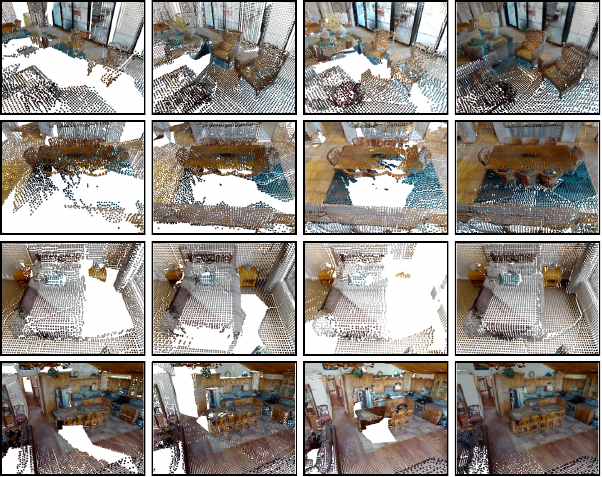}
    \caption{
    \small{
    \textbf{Qualitative Comparison. (Reconstruction quality).} Results comparing reconstruction quality of our method against baselines. By not just focusing on maximizing the coverage, but capturing all the structural details of indoor scenes, our method achieves the best visual fidelity. We are able to reconstruct \textit{completely} all elements that compose a house: not just chairs, dining tables and beds but floor and ceiling as well; while other methods are only able to achieve partial reconstruction. Notice the white areas that represent regions of the scene which these methods fail to scan and reconstruct.}}
    \label{fig:results_reconstruction}
    \vspace{-4mm}
\end{figure}

\vspace{-1mm}
\textbf{Baselines}. We benchmark the exploration efficiency of SCAGE against three established frameworks spanning both traditional and learning-based paradigms: our implementation of classic frontier-based method~\cite{classicalfrontier}, the sampling-based Next-Best-View Planner (NBVP)~\cite{schmid2020efficient}, 
and the recent state-of-the-art visual exploration approach, FrontierNet~\cite{frontiernet}. To guarantee a fair and rigorous comparison, we utilize the official baseline implementations and strictly adhere to the evaluation configurations defined by~\cite{frontiernet}, specifically testing across their identical subset of 10 validation meshes and enforcing their exact scene-specific maximum step budgets. We evaluate the baselines with perfect depth from the simulator and our method with both depth from simulator and Metric3D v2~\cite{metric3dv2}.

\textbf{Quantitative Evaluation}
We quantify exploration performance using two primary metrics: Volumetric Coverage and Reconstruction Quality (measured via Chamfer Distance) averaged over multiple runs from several starting positions. As summarized in Table~\ref{tab:quantitative_results}, \methodname{} consistently outperforms all baselines, achieving a maximum volumetric coverage of approximately 90\% across all evaluated scenes, a $\sim$15\% improvement over the best baseline FrontierNet~\cite{frontiernet}. This demonstrates our method's capacity to thoroughly navigate and map complex layouts without prematurely terminating or leaving significant regions unexplored. 

Crucially, this expansive coverage does not come at the expense of geometric detail. Our approach achieves the best overall reconstruction quality, as evidenced by yielding the lowest Chamfer Distance among all evaluated methods. As compared to the best baseline we achieve less than half the error using perfect depth and about 60\% using depth estimated by Metric3D v2~\cite{metric3dv2}). Furthermore, \methodname{} exhibits a remarkably low standard deviation across multiple evaluation runs with varying starting positions. This highlights the inherent robustness of the anomaly-driven formulation, proving that the system reliably captures high-fidelity maps regardless of the robot's initial initialization pose.

\textbf{Qualitative Results: Coverage and Exploration Trajectories.} To visually demonstrate the behavioral differences between the exploration strategies, we map the resulting trajectories and coverage across three distinct environments (Scenes 804, 812, and 824) in Fig.~\ref{fig:results_coverage}. Starting from identical initial locations, baseline methods frequently stall or leave substantial portions of the environment unexplored. 

Notably, while methods like FrontierNet~\cite{frontiernet} appear to achieve numerically competitive coverage, their qualitative behavior reveals a critical flaw. FrontierNet tends to maximize its volumetric reward by simply visiting the entrances of rooms, failing to penetrate deeper into the space to observe the structural elements that actually compose the room (see Fig. ~\ref{fig:results_coverage}). In contrast, by modeling exploration as a geometric anomaly minimization problem, \methodname{} actively seeks out unresolved geometries, driving the robot deep into rooms to capture fine details. 

\textbf{Qualitative Results: Reconstruction Quality.} This targeted resolution of structural anomalies translates directly into vastly superior visual fidelity, as highlighted in Fig.~\ref{fig:results_reconstruction}. While traditional methods typically yield fragmented or partial reconstructions of occluded objects, \methodname{} captures the complete structural details of the indoor scenes. Our anomaly-guided approach successfully and completely reconstructs a diverse array of environmental elements. This includes complex, cluttered objects such as chairs, dining tables, and beds, as well as broad, featureless surfaces like floors and ceilings. By prioritizing structural completeness rather than sheer spatial expansion, our method ensures the final 3D map is both volumetrically expansive and geometrically accurate.

\begin{table}[t]
\centering
\small
\setlength{\tabcolsep}{4pt}
\renewcommand{\arraystretch}{1.1}
\resizebox{0.98\linewidth}{!}{%
\begin{tabular}{l ccc c}
\toprule
& \multicolumn{3}{c}{Scene ID} \\
\cmidrule(lr){2-4}
Utility Function & 804 & 876 & 834 & Mean \\
\midrule
$\bar{s}_k$ & \cellcolor{second}$77.0 \pm 6.5$ & \cellcolor{second}$74.0 \pm 7.8$ & \cellcolor{second}$73.1 \pm 9.2$ & \cellcolor{second}$74.7$ \\

$\bar{s}_k / d(T_k)$ & $73.5 \pm 6.2$ & $69.9 \pm 8.6$ & $72.0 \pm 9.8$ & $71.8$ \\

$g(T_k) / d(T_k)$ & $38.1 \pm 6.2$ & $63.4 \pm 9.5$ & $63.1 \pm 9.1$ & $54.9$ \\

$g(T_k) \cdot \bar{s}_k / d(T_k)$ (Ours) & \cellcolor{first}$92.5 \pm 1.0$ & \cellcolor{first}$91.0 \pm 0.1$ & \cellcolor{first}$84.1 \pm 0.5$ & \cellcolor{first}$89.2$ \\
\bottomrule
\end{tabular}%
}
\caption{\small{\textbf{Ablation Study: Utility Function Design.} Volumetric coverage (\%) across three scenes comparing different utility formulations. $g(T_k)$ denotes volumetric information gain, $\bar{s}_k$ the mean cluster anomaly score, and $d(T_k) = \max(\|\mathcal{P}_{\text{robot}} - \mathcal{P}_{\text{cam},k}\|_2, \epsilon)$ the traversal cost. The results demonstrate that integrating all three terms is essential for maximizing exploration performance.}}
\label{tab:ablation_utility}
\end{table}

\textbf{Ablation Study: Utility Function Design.} We ablate the components of
our utility function (Eq.~\ref{eq:utility}) in Table~\ref{tab:ablation_utility}.
Relying solely on the anomaly score ($\bar{s}_k$) yields the second-highest
coverage, highlighting the raw exploratory power of targeting structural
ambiguities. Penalizing it by distance ($\bar{s}_k/d(T_k)$) degrades
performance, as the traversal term biases selection toward nearby targets and
causes the robot to over-commit locally. Dropping the anomaly signal entirely
($g(T_k)/d(T_k)$, equivalent to NBVP~\cite{schmid2020efficient}) is weakest of all: with no
structural cue to distinguish which unobserved regions are worth resolving, the
robot settles for whichever nearby free space is cheapest to reach, most visibly
in complex scenes such as 804. The terms are thus complementary, $\bar{s}_k$
identifying where geometry is unresolved and $g(T_k)$ pulling the robot toward
large unobserved volumes, counteracting the local bias of $d(T_k)$. Only their
combination unlocks peak exploration behavior.

\begin{figure}[t]
    \centering
    \includegraphics[width=1\linewidth]{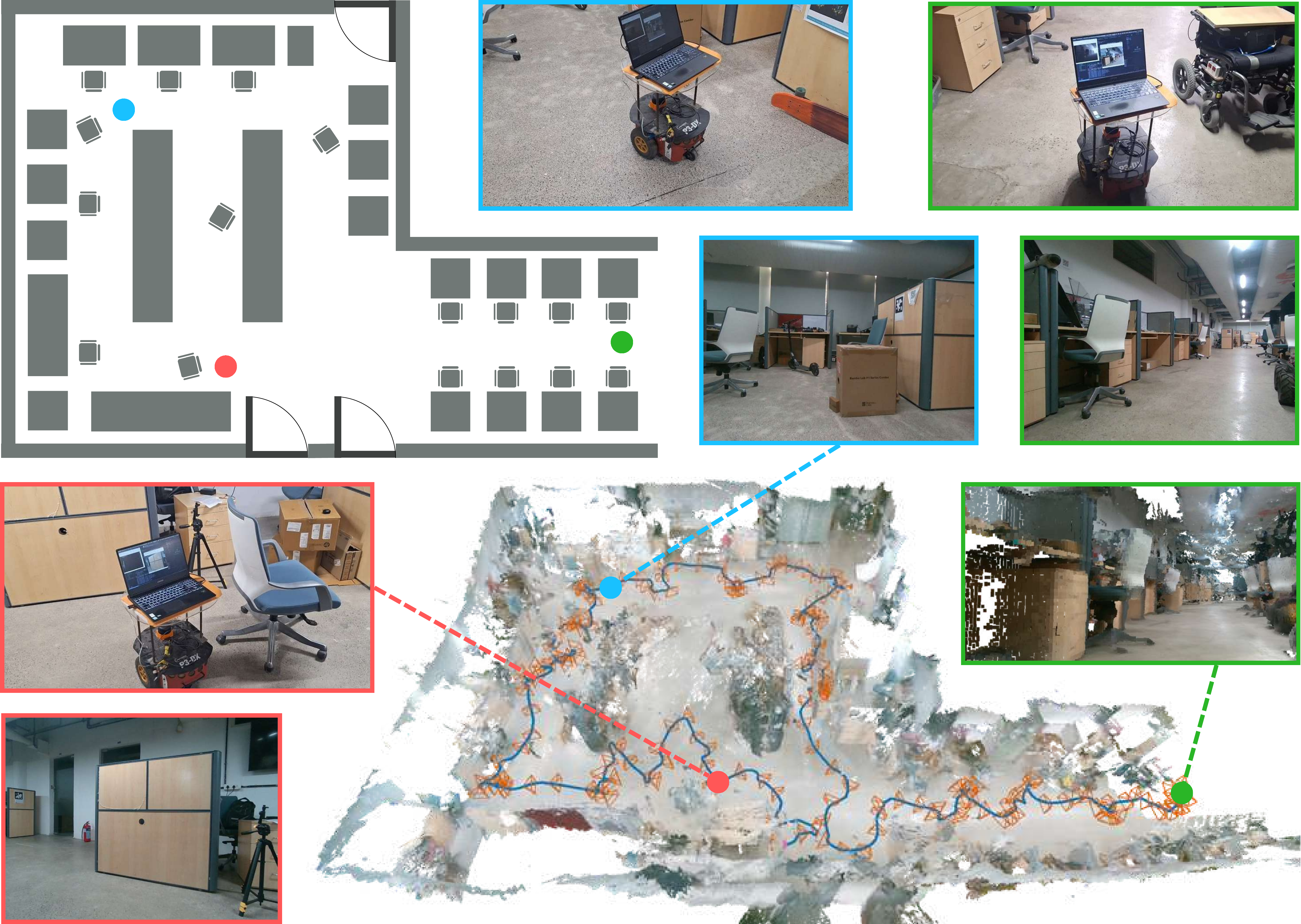}
    \caption{
    \small{
    \textbf{Real world validation}. Schematic representative map of the lab (top-left) and the resulting 3D point cloud reconstruction (roof cropped for visibility). The robot's trajectory begins at the green starting point. The surrounding color-coded images corresponding to the green, blue, and red markers, display the robot in the environment alongside its egocentric view at that exact moment during capture, including a point cloud from the starting location.}
    }
    \label{fig:real}
    \vspace{-4mm}
\end{figure}

\vspace{-1mm}
\subsection{Real-world Validation}

We implement \methodname{} as a ROS package and deploy it on a P3DX robot equipped with a RealSense D455 depth camera, providing 640$\times$360 RGB-D images at 10 Hz. All computations are performed on a laptop with an i5-9300H CPU, 32 GB of RAM, and an RTX 2060 GPU, achieving $\sim$7 Hz inference for real-time point cloud processing. We show the results in Fig.~\ref{fig:real}. Despite being trained on perfect-looking houses from the Matterport3D dataset, \methodname{} is robust in the real world environment, with the robot successfully mapping the structural components of the large indoor scene. Notably,~\networkname{} flags incomplete chairs, cubicle partitions, and walls in our lab as anomalies despite these specific instances differing from those seen during training, confirming that the network has learned the underlying geometry of what constitutes such structures rather than memorizing training exemplars.
However object categories absent from the training distribution, such as other mobile robots in the lab, elicit weak anomaly responses and are observed via anomalies of the surrounding area.

\begin{figure}[t]
    \centering
    \includegraphics[width=0.98\linewidth]{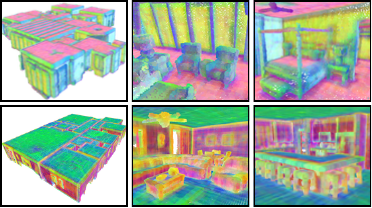}
    \caption{
    \small{
    \textbf{Emergent Structural Understanding.} PCA visualization of features extracted from the \networkname{} encoder on scenes 876 (top) and 879 (bottom). Structurally similar regions share consistent feature representations (colors) across different rooms. For example, planar horizontal surfaces like floors, ceilings, and beds exhibit uniform color, while objects such as fans, chairs, and kitchen cabinetry stand out in sharp contrast against their structural background.}}
    \label{fig:ablation_semantic}
    \vspace{-5mm}

\end{figure}

\subsection{Emergent Structural Understanding}
Beyond quantitative coverage and reconstruction metrics, we observe that the success of \methodname{} is rooted in the network's deep, intrinsic understanding of indoor environments. To investigate what the network learns, we extract the dense latent features from the \networkname{} encoder and project them into RGB space using Principal Component Analysis (PCA) (Fig.~\ref{fig:ablation_semantic}). Even though the network is trained entirely from scratch without any explicit semantic supervision, it naturally learns to group similar structural and semantic elements. Horizontal surfaces (floors, ceilings, and beds) share consistent feature distributions, while distinct architectural objects like ceiling fans, chairs, and tables are highly isolated. This emergent structural awareness highlights the remarkable expressivity of the Point Transformer V3~\cite{ptv3} backbone. Because the model genuinely learns the fundamental building blocks of how houses are constructed, it can accurately identify \textit{scene anomalies} to drive exploration. Furthermore, the richness of these learned representations suggests that our pre-trained encoder could serve as a powerful, off-the-shelf feature extractor for complex downstream applications, such as zero-shot semantic segmentation or object goal navigation.

\section{CONCLUSION}
\label{sec:conclusion}

We presented \methodname{}, a novel autonomous exploration framework that successfully bridges the gap between spatial coverage and high-fidelity 3D reconstruction. By reframing exploration as a geometric anomaly-minimization problem, we move beyond traditional heuristics for exploration. At the core of our approach is \networkname{}, a 3D denoising autoencoder that learns the foundational structural priors of typical indoor environments directly from localized point cloud chunks. 

During navigation, our system continuously evaluates live 3D observations against these learned expectations, identifying geometric contradictions as \textit{scene anomalies}. By dynamically targeting and resolving these structural ambiguities from optimal vantage points, \methodname{} seamlessly couples spatial discovery with precise geometric mapping. Extensive evaluations demonstrate that our approach achieves superior volumetric coverage ($\sim$90\% across scenes, $\sim$15\% improvement) and higher reconstruction quality (half the error) compared to state-of-the-art baselines. Ultimately, by shifting the primary exploratory signal from simply finding unmapped voids to actively achieving structural comprehension, this work establishes a new direction for autonomous 3D scene exploration.

\textbf{Acknowledgements.} The authors acknowledge the support provided by MeitY,
Govt. of India, under the project ``Capacity Building for
Human Resource Development in Unmanned Aircraft System
(Drone and Related Technology)."

\addtolength{\textheight}{-12cm}   %

\bibliographystyle{IEEEtran}
\bibliography{references}

\end{document}